\pdfoutput=1
\documentclass[11pt]{article}

\usepackage[]{EMNLP2022}

\usepackage{times}
\usepackage{latexsym}
\usepackage{multirow}   %  table
\usepackage{graphicx}   %  .png
\usepackage{enumitem}
\usepackage[linesnumbered,ruled,vlined]{algorithm2e} % \algorithm
\usepackage{amsmath}    %  symbols
\usepackage{amssymb}    %  symbols
\usepackage{latexsym}
\usepackage[inkscapelatex=false]{svg}
\usepackage{url}    % \url
\usepackage{colortbl, booktabs} % To thicken table lines
\usepackage{makecell} % automatic words wrap
\usepackage{color}  %  color
\usepackage{array}
\usepackage{hyperref}   %  \href{url}{caption}
\usepackage[all]{nowidow}   % prevents the appearance of an ugly single line on the second page.
\usepackage[subtle]{savetrees} % The goal of the savetrees package is to pack as much text as possible onto each page

\SetKwInput{KwInput}{Input}                % Set the Input
\SetKwInput{KwOutput}{Output}              % set the Output

\usepackage[T1]{fontenc}
\usepackage[utf8]{inputenc}

\usepackage{microtype}

\usepackage{inconsolata}

\title{EtriCA: Event-Triggered Context-Aware Story Generation \\ Augmented by Cross Attention}

\author{Chen Tang\textsuperscript{1}, Chenghua Lin\textsuperscript{2}\footnotemark[1] , Henglin Huang\textsuperscript{1}, Frank Guerin\textsuperscript{1} ~and  Zhihao Zhang\textsuperscript{3} \\
\textsuperscript{1}Department of Computer Science, The University of Surrey, UK \\
\textsuperscript{2}Department of Computer Science, The University of Sheffield, UK \\
\textsuperscript{3}School of Economics and Management, Beihang University, Beijing, China \\
\texttt{\{chen.tang,hh01034,f.guerin\}@surrey.ac.uk}  \\
\texttt{c.lin@sheffield.ac.uk},  \texttt{zhhzhang@buaa.edu.cn}}

\begin{document}
\maketitle

%  for authors' footnotes
\renewcommand{\thefootnote}{\fnsymbol{footnote}} 
\footnotetext[1]{Corresponding author.} 
\renewcommand{\thefootnote}{\arabic{footnote}} 

\begin{abstract}

One of the key challenges of automatic story generation is how to generate a long narrative that can maintain fluency, relevance, and coherence. Despite recent progress, 
current story generation systems still face the challenge of how to effectively capture contextual and event features, which has a profound impact on a model's generation performance. To address these challenges, we present EtriCA, a novel neural generation model, which improves the relevance and coherence of the generated stories through residually mapping context features to event sequences with a cross-attention mechanism. Such a feature capturing mechanism allows our model to better exploit the logical relatedness between events when generating stories. Extensive experiments based on both automatic and human evaluations show that our model significantly outperforms state-of-the-art baselines, demonstrating the effectiveness of our model in leveraging  context and event features.

\end{abstract}

% =============================== Section 1 ==================================
\section{Introduction}
Story Generation aims to generate fluent, relevant and coherent narratives conditioned on a given context. 
As the task is notoriously difficult, a common strategy is to employ storylines composed of events to support the generation process~\cite{yao2019plan, chen-etal-2021-graphplan, alhussain2021automatic, tang2022ngep}. This process imitates the behavior of human writers. Firstly, a story will start from a sketch of key words containing events, and then human writers will unfold the story following the track of planned event sequences. 

% ----------- fig: examples -----------
\begin{figure}[tb]
\centering
\includegraphics[width=\columnwidth]{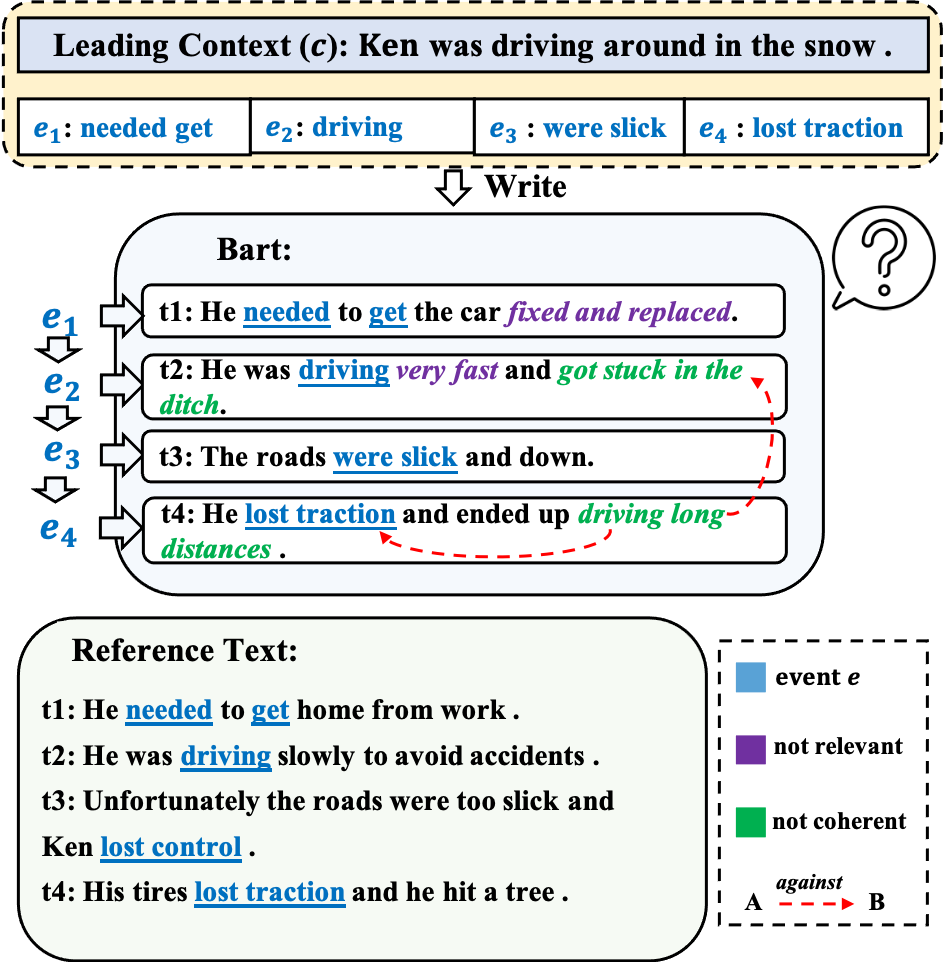}
\caption{Conditioned on leading context and reference events (extracted from reference stories), existing generation models still suffer from problems of relevance and coherence. For instance, we fine-tune BART \cite{lewis-etal-2020-bart} to generate stories. The leading context and reference text in this example are collected from ROC Stories \cite{mostafazadeh-etal-2016-corpus}. Some conflicts among them are observed and coloured. 
} 
\label{fig:examples}
\end{figure}
% ----------- end of fig -----------

Despite recent progress, existing approaches are still ineffective in exploiting planned events when generating stories. 
Usually, pre-trained generation models, e.g., BART~\cite{goldfarb-tarrant-etal-2020-content,clark-smith-2021-choose,huang2022improving} are employed to generate stories after event planning. However, as shown by the conflicts in Figure \ref{fig:examples}, the separate sentences generated by BART look reasonable, but there are several issues observed considering the whole story: As a commonsense story, if the car needs to be \textit{``fixed and replaced''} then it is too broken to \textit{``drive around''}; \textit{``Ken''} should not drive the car \textit{``very fast''} in the \textit{``snow''}; If \textit{``Ken''} \textit{``got stuck in the ditch''} or \textit{``lost traction''}, he cannot then be \textit{"driving long distances"}. We hypothesise that these problems come from the inadequacy of capturing contextual features when keeping track of event sequences, because (\romannumeral1) the planned events generally lack background information, e.g., Ken (the character) and snow (the scene) and (\romannumeral2) training stories may have the same events but different reference stories, which may lead to confusion during inference if not considering the story-specific scenario. 

Therefore, to address these challenges we propose \textbf{EtriCA} - a novel \textbf{E}vent-\textbf{Tri}ggered \textbf{C}ontext-\textbf{A}ware end-to-end framework for story generation.
Given both leading context and planned events, EtriCA  can more effectively capture contextual and event features from inputs than state-of-the-art (abbr. SOTA) baseline models. Traditional generation models struggle to learn contextual representations when implicitly keeping track of the state of events due to feature differences of events and contexts. As an abstract storyline, an event sequence only contains schematic information related to actions (e.g. the verb), while the context usually records story-specific details, e.g., the scene and characters in a story. 

\begin{figure*}[tb]
\centering
% \includesvg[width=1.3\columnwidth]{figures/mechanism.svg}
\includegraphics[width=1.3\columnwidth]{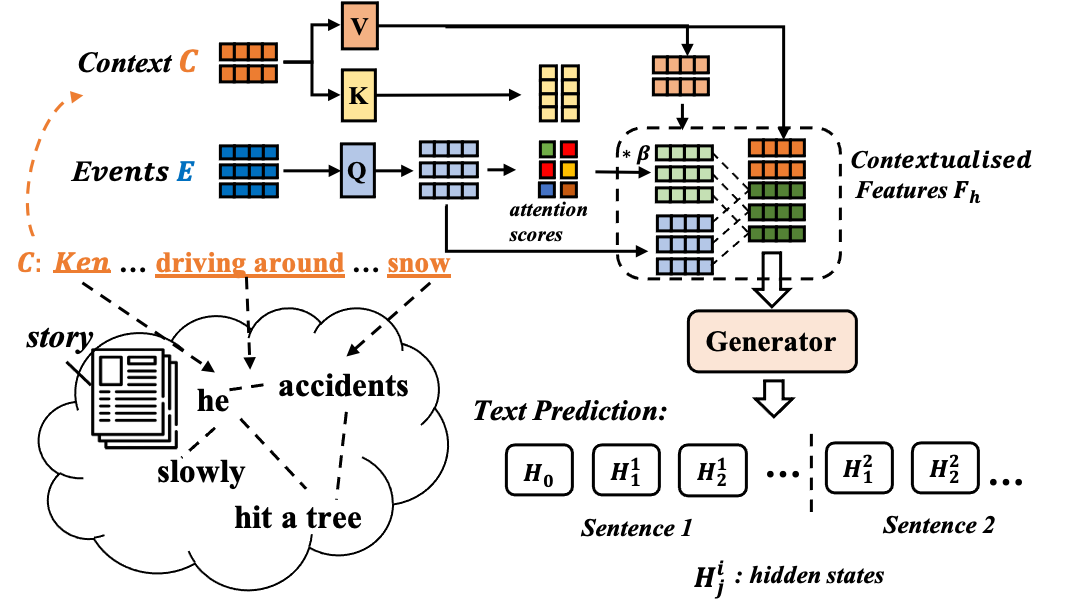}
\caption{The overview of the event feature contextualising process. The leading context coloured in red contains some important information which affects the generation process, e.g., the weather "snows" may lead to "accident". These implicit clues help the neural generator to disambiguate the context of events. We firstly fuse both context and event features, and then feed them to the generator.}
\label{fig:mechanism}
\end{figure*}

To comprehensively leverage both features, we draw inspiration from prior work dealing with information fusion \cite{chen-etal-2018-hybrid,xing-etal-2020-automatic,he-etal-2020-scene,you-etal-2020-hard, wang-etal-2021-fast,tang2022recent}
to encode heterogeneous features with a cross attention mechanism \cite{gheini2021cross}. We aim to inform our model of the context background when the neural module unfolds each event into a narrative. We propose a novel neural module that learns to implicitly map contextual features to event features through information fusion on their numeric vector spaces (we call this process contextualising events). The whole process is illustrated in Figure.~\ref{fig:mechanism}. With the contextualised event features, an autoregressive decoder is employed to dynamically generate stories by learning to unfold the contextualised events. We also introduce an auxiliary task of Sentence Similarity Prediction \citep{guan-etal-2021-long} to enhance the coherence between event-driven sentences. 

To support research on event-driven story generation, we propose a new task formulated by writing stories according to a given leading context and event sequence. We improve the event extraction framework of \citet{chen-etal-2021-graphplan} by exploiting dependency parsing to capture event related roles from sentences, instead of using heuristic rules. We also present two datasets where multi-sentence narratives from existing datasets are paired with event sequences using our automatic event extraction framework. Importantly, our task formulation can also benefit the study of controllable story generation, considering there is increasing interest in storyline-based neural generative frameworks \cite{xu-etal-2020-megatron, ghazarian-etal-2021-plot, chen-etal-2021-graphplan}.  According to our extensive experiments, EtriCA performs better than baseline models considering the metrics of fluency, coherence, and relevance. Our contributions\footnote{The related code is available at \url{https://github.com/tangg555/EtriCA-storygeneration}} can be summarised as follows:

\begin{itemize}[noitemsep,nolistsep,leftmargin=*]
    \item A new task formulation for event-driven story writing, which requires the generation model to write stories according to a given leading context and event sequence. 
    \item We annotate event sequences on two existing popular datasets for our new task, and introduce new automatic metrics based on semantic embeddings to measure the coherence and relevance of the generated stories.
    \item We propose a neural generation model \textbf{EtriCA}, which leverages the context and event sequence with an enhanced cross-attention based feature capturing mechanism and sentence-level representation learning.
    \item We conduct a range of experiments to demonstrate the advances of our proposed approach, and comprehensively analyse the underlying characteristics contributing to writing a more fluent, relevant, and coherent story. 
\end{itemize} 

% ----------- fig:mechanism -----------

% ----------- fig:dep -----------
\begin{figure*}[htp]
\centering
\includegraphics[width=1.8\columnwidth]{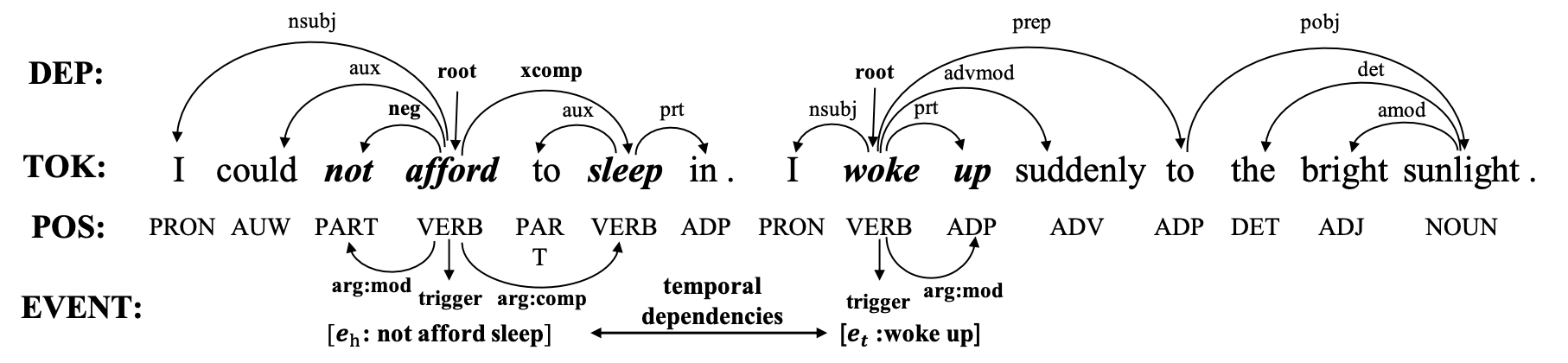}
\caption{An example to illustrate the process of event extraction. \textit{TOK} is the basic unit of a sentence. \textit{POS} is the part of speech, and \textit{DEP} stands for dependencies between tokens. Through parsing dependencies, the event trigger (also recognised as the root of sentence) filters all significant roles to represent a complete action. Meanwhile, extracted neighbour events are considered to have temporal relations.}
\label{fig:dep}
\end{figure*}
% ----------- end of fig -----------
% =============================== Section 2 ==================================
\section{Related Work}

\subsection{Neural Story Generation} 
Before the surge of deep learning techniques, story generation models only generated simple sentences and heavily relied on manual designs \cite{mcintyre-lapata-2009-learning,woodsend-lapata-2010-automatic,mcintyre-lapata-2010-plot,huang-huang-2013-optimized,kybartas2016survey}. Since neural story generation came into being, end-to-end neural models, especially pre-trained models, e.g., BART \cite{lewis-etal-2020-bart} and GPT-2 \cite{radford2019language}, are widely employed as the main module of story writing \cite{rashkin-etal-2020-plotmachines, guan-etal-2020-knowledge,goldfarb-tarrant-etal-2020-content,clark-smith-2021-choose}. However, it is hard to guarantee logical correctness for naive Seq2Seq models when the generated text is growing longer, so recent work is exploring multi-step generations which implement neural models in traditional generative pipelines \cite{guan-etal-2021-long}. For example, \citet{yao2019plan,goldfarb-tarrant-etal-2020-content,chen-etal-2021-graphplan} split story generation into planning (inputs to events) and writing (events to stories), and leverage two neural generation models to learn them.

\subsection{Event Planning for Story Generation} \label{sec:planning}
At the planning stage, prior research \cite{yao2019plan, rashkin-etal-2020-plotmachines, goldfarb-tarrant-etal-2020-content,jhamtani-berg-kirkpatrick-2020-narrative, ghazarian-etal-2021-plot} mostly focused on extracting event sequences from the reference text as the ground truths of plot planning, and then leveraged neural models \cite{radford2019language,lewis-etal-2020-bart} to predict events with given leading context or titles. Events have a lot of representation formats, e.g., verbs, tuples, key words, etc. Among them a straight forward approach is extracting verbs as events \cite{jhamtani-berg-kirkpatrick-2020-narrative, guan-etal-2020-knowledge, kong2021stylized}, which is also the method we followed. However,  verbs alone are not good enough to keep information integrity. For instance, semantic roles like negation (not) are significant for correct understanding. \citet{peng-roth-2016-two} and \citet{chen-etal-2021-graphplan} use some heuristic rules to include these semantic roles, but those heuristic rules are not complete to include all the key roles. Therefore, inspired by related works \cite{rusu-etal-2014-unsupervised,bjorne-salakoski-2018-biomedical,huang-etal-2018-zero} in open-domain event extraction, we 
propose an event extraction workflow based dependency parsing to capture essential components for verb phrases in sentences as events.

% =============================== Section 3 ==================================
\section{Methodology} \label{sec:methodology}

% ######################### Section 3.1 ########################
\subsection{Task Formulation} \label{sec:task}
 Under the umbrella of controllable story generation we define the following task: 
 write a story that leverages both the given leading context and a given planned event sequence.
 Our primary goal is to investigate how to consider the  context while keeping track of the given event sequence with neural generation models, so we expand the original context-aware story generation settings of \citet{guan-etal-2021-long} by adding an event sequence, for each leading context, as the storyline to follow. 

\noindent\textbf{{Input:}}~The input for each sample includes a leading context $ C = \{c_1, c_2, ..., c_n\} $ which acts as the first sentence of a story, and an event sequence $ E = \{e_1, e_2, ..., e_m\}$ as a storyline to build up a sketch for a story. $c_i$ means the $ i $-th token of the leading context, and $ e_i $ means the $ i $-th event representing the $ i $-th sentence in a story.

\noindent\textbf{{Output:}}~The output is a multi-sentence story $ S = \{s_1^1, s_2^1, ..., s_1^2 ..., s_n^m\}$. , $ s_j^i $ denotes the $j$-th token of $ i $-th sentence in a story.

% ----------- fig:arch -----------
\begin{figure*}[htb]
\centering
\includegraphics[width=2.1\columnwidth]{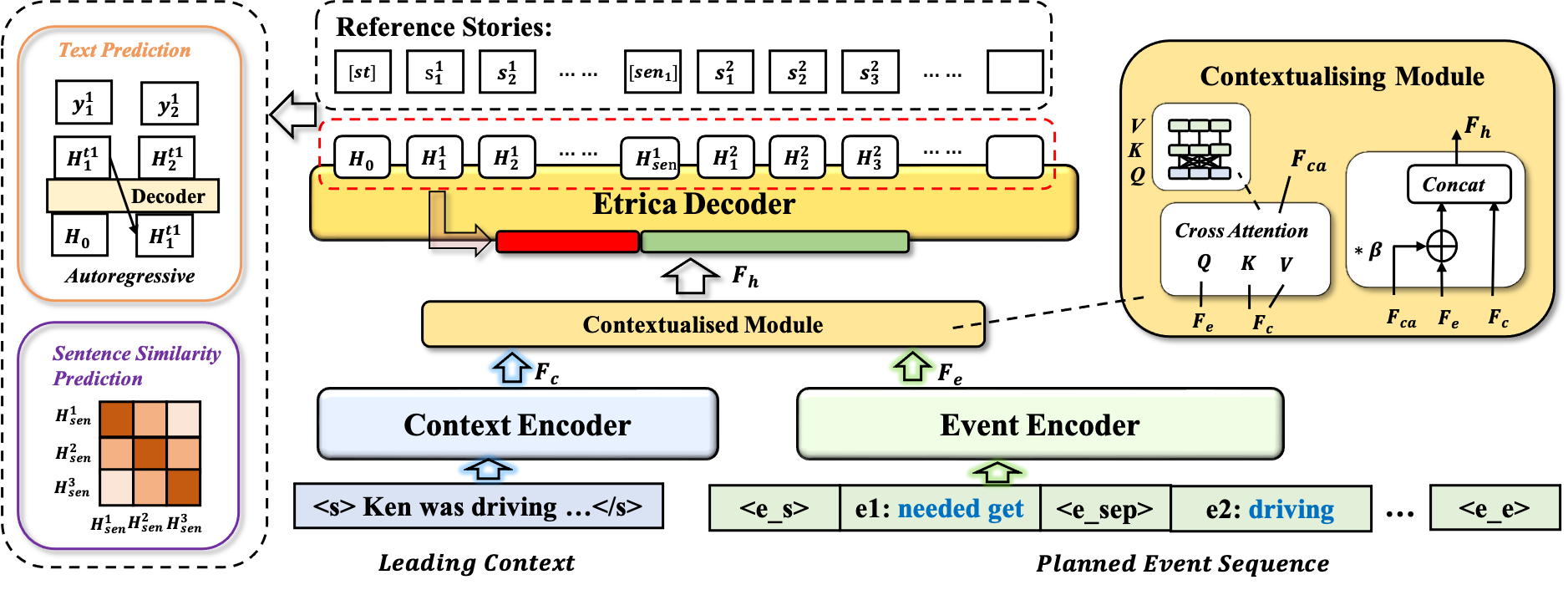}
\caption{The overview of EtriCA architecture. The technique details are explained in Sec~\ref{neural-model}. When training, in addition to predicting text tokens $\{y_1^1, ..., y_j^i\}$ one by one, we train the decoder to learn sentence-level representations by the similarity prediction auxiliary task shown in the dotted box.  Through representation learning, neural models learn how to generate the reference-like stories with given leading context and planned event sequence.}
\label{fig:model}
\end{figure*}
% ----------- end of fig -----------

% ######################### Section 3.2 ########################
\subsection{Event Sequence Preparation} \label{event extraction}

Following the prior work of event extraction \cite{chen-etal-2021-graphplan} (discussed in Sec.~\ref{sec:planning}), we present an automatic framework which includes all verb related roles by analysing dependencies between words\footnote{We use $spaCy$ (\url{https://spacy.io/}) to split sentences and parse dependencies between words in each sentence. }. The event representation is not the focus of this study however. Figure \ref{fig:dep} shows an example of the event extraction process. The details of the event schema are appended in Appendix.~\ref{apx:schema}.

% ######################### Section 3.3 ########################
\subsection{Neural Framework} \label{neural-model}

At the writing stage, conditioned on a leading context $C$ and planned events $E$ (see Sec.~\ref{sec:task}), the neural model generates a multi-sentence story $S$. Figure~\ref{fig:model} shows the overview of the whole process.

\noindent\textbf{{Planned Events Representation}}~
The language models of conventional generation frameworks either based on transformers \cite{vaswani2017attention} or RNNs \cite{ghosh-etal-2017-affect}, are commonly designed to encode natural language input  text, so the extracted events need to be firstly serialised to plain text. We separate the string format of events introduced in Sec.~\ref{event extraction} with special tokens, e.g., ``<e\_s> needed get <e\_sep> ... <e\_e>'', where ``<e\_s>'',``<e\_sep>'', ``<e\_e>'' represent the start, separation, and end of planning, respectively. 

\noindent\textbf{{Contextualised Features Representation}}~
At the encoding stage, neural models have the inputs of $C$ and $E$ which have feature differences as mentioned above. Conventional end-to-end models usually concatenate embeddings of different inputs, since neural encoders  capture their features in numeric vector space (e.g. through self-attention).  However, when the event sequence grows longer, the growing concatenated embeddings may decrease the influence of $C$ (we discuss this in Appendix~\ref{apx:contextualised}). Instead, we firstly leverage two separate BART~\cite{lewis-etal-2020-bart} encoders to incorporate features, and then fuse features with multi-head attentions calculated as below:  
% ----------- equation-----------
\begin{align}
    & F_c = \text{Encoder}_c(C); F_e = \text{Encoder}_e(E) \\
    & Q_i = {W_i^Q} {F_e}, K_i = {W_i^K} {F_c}, V_i = {W_i^V} {F_c},\\
    & A_i = \text{softmax}(\frac{Q_iK_i^\mathrm{T}}{\sqrt{d_k}})V_i \\ 
    & F_{ca} = \text{Concat}(A_1, ..., A_m)W^M
\end{align}
% ----------- end of equation -----------
where $Encoder_c$ and $Encoder_e$ inherit pre-trained parameters from BART but do not share trainable parameters when fine-tuning. $F_{c}$ and $F_{e}$ stand for the features captured from $C$ and $E$, respectively. $i$ denotes the $i$-th head of the attention scores which have $m$ heads in total. $W_i^Q$, $W_i^K$, $W_i^V$, and $W^M$ are trainable parameters. The $i$-th head attention $A_i$ is the attention-based weight sum of the feature matrix. Finally, the obtained $F_{ca}$ represents the attentions of ongoing events under the consideration of context.

In order to contextualise the input event features, we incrementally add the $F_{ca}$ to the original event features $F_{e}$ so that neural models are forced to learn the context gap between event sequences and stories, i.e.,
% ----------- equation-----------
\begin{align}
    & F_{he} = F_e + \beta \odot F_{ca} \label{eq:beta} \\
    & F_h = \text{Concat}(F_c, F_{he})
\end{align}
% ----------- end of equation -----------
where $\beta$ denotes the scale factor of $F_{ca}$. $\beta \odot F_{ca}$ is the representation of the context gap trained via residual mapping. $F_h$ concatenates both leading context features $F_c$ and contextualised event features $F_{he}$. These are fed into a neural decoder to predict tokens and sentence representations.

\paragraph{Decoding and Sentence-level Fitting}
As in other conventional generation systems, we employ an auto-regressive decoder to generate story tokens $y_t$ as equations below.
% ----------- equation-----------
\begin{align}
    & H_t = \text{Decoder}(y_{<t}, F_h) \\
    & P(y_t|y_{<t}, X) = \text{softmax}(H_tW) \\
    & y_t \stackrel{sampling}{\longleftarrow} P(y_t|y_{<t}, F_h)
\end{align}
% ----------- end of equation -----------
where $t$ denotes a time step. $X$ denotes the input to the neural model. $H_t$ is the $t$-th hidden state of the decoder module. $H_t$ is computed from the information of both context and events contained in $F_h$ and the prior predicted story $y_{<t}$. $W$ is a trainable parameter, and $P(y_t|y_{<t}, F_h)$ is the probability distribution of the vocabulary including special tokens. Through a sampling strategy, e.g., $\mathit{argmax}$, we collect the predicted token $y_t$. 

In addition to the token-level representation, we introduce an auxiliary task of Sentence Similarity Prediction \citep{guan-etal-2021-long} to learn sentence-level representations and training methods. Due to the page limit, the details are moved to  Appendix~\ref{apx:methodology}.

\paragraph{Training and Inference}
As shown in Figure~\ref{fig:model}, the neural model is trained to fit both token and sentence level references as follows:
% ----------- equation-----------
\begin{align}
    & \mathcal{L}_{lm} = - \frac{1}{N}\sum_{t=1}^N logP(y_t|y_{<t}, X) \\
    & \mathcal{L}_{sent} = \frac{1}{m^2} \sum_{i=1}^m \sum_{j=1}^m (max|sim_{ij}^{s} - sim_{ij}^{y}|, \Delta) \label{eq:delta}\\
    & \mathcal{L}_{overall} = \mathcal{L}_{lm} + \lambda \mathcal{L}_{sent} \label{eq:lam}
\end{align}
% ----------- end of equation -----------
where $\mathcal{L}_{lm}$ is the cross-entropy loss of $P(y_t|y_{<t}, F_h)$. $\mathcal{L}_{sent}$ is the loss of predicted sentence similarities. $sim_{ij}^{s}$ and $sim_{ij}^{y}$ denote the sentence similarities between the $i$-th and $j$-th sentences in a reference story and a generated stories, respectively.  $\lambda$ is an adjustable scale factor, and $\mathcal{L}_{\mathit{overall}}$ is the overall loss. By minimising  $\mathcal{L}_{\mathit{overall}}$, the neural model learns to predict a human-like story.
The Sentence Similarity Prediction only works on training, and when doing inference the neural model finally outputs stories without those special tokens.

% =============================== Section 4 ==================================
% ----------- tab: referenced metrics -----------
\begin{table*}[ht]
\centering \small
%\resizebox{0.9\linewidth}{!}{
\begin{tabular}{l|cccccc|cccccc}
\toprule[1pt]
\multirow{2}{*}{\textbf{Models}} & \multicolumn{5}{c}{\textbf{ROC Stories}} & \multicolumn{5}{c}{\textbf{Writing Prompts}} \\
  & \textbf{PPL}$ \downarrow $ & \textbf{R-1}$ \uparrow $ & \textbf{R-2}$ \uparrow $ & \textbf{R-L}$ \uparrow $ & \textbf{B-1}$ \uparrow $ & \textbf{B-2}$ \uparrow $   
  & \textbf{PPL}$ \downarrow $ & \textbf{R-1}$ \uparrow $ & \textbf{R-2}$ \uparrow $ & \textbf{R-L}$ \uparrow $ & \textbf{B-1}$ \uparrow $ & \textbf{B-2}$ \uparrow $   \\
\midrule
\textbf{P\&W$_{l+e}$} & 6.22 & 22.82 & 2.65 & 15.90 & 0.297 & 0.150      & 16.47 & 23.49 & 1.74 & 12.17 & 0.259 & 0.086  \\
\textbf{GPT-2$_{l+e}$} & 10.02 & 29.85 & 6.45 & 20.58 & 0.347 & 0.201    & 49.48 & 18.59 & 2.18 & 10.38 & 0.130 & 0.051  \\
\textbf{BART$_{l+e}$}  & 3.39 & 48.74 & 21.95 & 40.69 & 0.505 & 0.351     & 10.84 & 37.19 & 8.14 & 22.73 & 0.351 & 0.174 \\

\textbf{HINT$_{l+e}$}  & 3.97 & 46.71 & 20.81 & 37.21 & 0.488 & 0.337      & 14.45 & 38.86 & 8.98 & 23.06 & 0.373 & 0.190  \\
\midrule
\textbf{EtriCA (ours)} & \textbf{2.88} & \textbf{49.29} & \textbf{22.59} & \textbf{41.43} & 0.506 & 0.354   & \textbf{8.11} & \textbf{39.90} & \textbf{9.65} & \textbf{25.21} & \textbf{0.387} & \textbf{0.202}  \\ 
\textbf{- w/o sen} & 3.33 & 49.18 & 22.39 & 41.09 & \textbf{0.512} & \textbf{0.359}      & 9.88 & 39.88 & 9.37 & 24.86 & 0.385 & 0.199 \\
\textbf{- w/o cm} & 2.97 & 48.53 & 21.55 & 40.34 & 0.499 & 0.345    & 9.15 & 36.08 & 7.55 & 21.01 & 0.356 & 0.175 \\ 
\textbf{- w/o leading} & 3.24 & 42.55 & 17.21 & 35.90 & 0.450 & 0.287     & 9.37 & 35.46 & 7.22 & 20.69 & 0.357 & 0.172 \\ 
\textbf{- w/o events} & 4.50 & 24.51 & 2.70 & 16.86 & 0.311 & 0.156    & 12.77 & 23.77 & 1.89 & 12.26 & 0.263 & 0.089 \\
\bottomrule[1pt]
\end{tabular}
%}
\caption{Automatic evaluation of referenced metrics on ROC and WP datasets. 
The best performance in each line is highlighted in \textbf{bold}. $ \uparrow / \downarrow $ means the higher/lower the better, respectively. $_{l+e}$ means the input of the model concatenates the leading context and event sequence. 
\textbf{w/o sen}, \textbf{w/o cm}, \textbf{w/o leading}, and \textbf{w/o events} means  ablating the auxiliary task of sentence similarity prediction, respectively, the contextualising module, the leading features, and the event features.}
\label{tab:referenced}
\end{table*}
% ----------- end of tab-----------

\section{Experiment}
% ######################### Section 4.1 ########################
\subsection{Datasets}
In this study, we annotate two popular datasets, ROCStories (ROC) \cite{mostafazadeh-etal-2016-corpus} and Writing Prompts (WP) \cite{fan-etal-2018-hierarchical} with extra event sequences as our benchmarks. We follow the settings of prior work \cite{xu-etal-2020-megatron,guan-etal-2021-long} to preprocess these data. The stories in both datasets are split into sentences by NLTK \cite{bird-loper-2004-nltk}. The data of ROC are delexicalised by masking all the names with tokens of \textit{[MALE]}, \textit{[FEMALE]}, and \textit{[NEUTRAL]}. The data of WP are recollected from the original development and test set, and we retain the first eleven sentences in each story\footnote{The original WP dataset is too large, and the topics are unconstrained.}. For both datasets, the first sentence in a story is extracted to be the leading context $C$ as the input, and the rest is used as the reference story $S$. Finally, we obtain a long story dataset, WP (10 sentences), and a short story dataset, ROC (4 sentences) for the following experiments. The event sequence $E$ is extracted from the reference story $S$ as the planned plot to guide story generation. Statistically, the ROC has stories as the Train/Dev/Test set of 88344/4908/4909 stories, respectively, and the split of WP is 26758/2000/2000.

% ######################### Section 4.2 ########################
\subsection{Baselines}
%\textcolor{orange}{Current Story Generation tasks have slightly different settings, but they are usually based on several neural generation models.} 
We compare EtriCA with following SOTA generation models: (1) \textbf{P\&W} (Plan and Write) \cite{yao2019plan}: The main architecture is based on a BiLSTM with an attention mechanism \cite{garg-etal-2019-jointly}. To make the comparison more fair, we enhance the original code by replacing original static word embeddings with the dynamic embeddings of the pre-trained BART; (2) \textbf{GPT-2} \cite{radford2019language}: A popular auto-regressive generative model which has been widely used in prior works \cite{rashkin-etal-2020-plotmachines, guan-etal-2020-knowledge,clark-smith-2021-choose}; (3) \textbf{BART}: This is a composed model constructed with a BERT-like \cite{devlin-etal-2019-bert} encoder and a GPT-like decoder, and shown advances in prior NLG works \cite{goldfarb-tarrant-etal-2020-content,clark-smith-2021-choose}. (4) \textbf{HINT} \cite{guan-etal-2021-long}: It is currently the SOTA framework on context-aware story generation, which enhanced the coherence and relevance through training with two training objectives.

% ######################### Section 4.3 ########################
\subsection{Implementation Details}
The main contribution of our generation model is the contextualising module, which can adapt to other encoder-decoder frameworks. Therefore, we employ the encoders and decoders from the BART framework \cite{lewis-etal-2020-bart}, which has shown strong performance in prior studies \cite{goldfarb-tarrant-etal-2020-content,guan-etal-2021-long}, to build our neural generation model. 
We fine-tune our generation model based on a publicly available BART checkpoint\footnote{The checkpoint of bart-base loaded from \url{https://huggingface.co/facebook/bart-base}} and  fix the random seed to 42.
All of our code is implemented in PyTorch, and trained with the PyTorch Lightning framework. More details of the hyper-parameters of the model, training and inference are described in Appendix~\ref{apx:implementation}.

% ######################### Section 4.4 ########################
\subsection{Automatic Evaluation}
% ---- 4.4.1
\subsubsection{Evaluation Metrics}
\textbf{Perplexity (PPL)} measures the uncertainty of generated tokens predicted by neural models. \textbf{ROUGE-n (R-n)} \cite{lin-2004-rouge} is a set of reference metrics measuring the coverage rate between generated stories and the referenced stories where $n$ denotes n-grams. \textbf{BLEU-n (B-n)} \cite{papineni-etal-2002-bleu} is also a set of reference metrics to compute n-gram overlaps between the generated stories and the references.
\textbf{Lexical Repetition-n (LR-n)} is an unreferenced metric to compute the percentage of generated stories which have a 4-gram repeated at least $n$ times \cite{shao-etal-2019-long}. 
\textbf{Distinction-n (D-n)} is an unreferenced metric qualifying the distinction of stories by measuring the ratio of distinct n-grams to all those generated n-grams \cite{li-etal-2016-diversity}. \textbf{Intra-story Repetition} \citep{yao2019plan} measures the repetition of each sentence in a story by the overlaps of trigrams. \textbf{Intra-story Coherence and Relevance} \cite{xu-etal-2018-better},  originally used in dialogue evaluation, is based on cosine similarity between semantic embeddings\footnote{Glove Vectors are used here. \url{https://nlp.stanford.edu/projects/glove/}} to calculate sentence-level coherence and relevance. This approach is used to measure the relatedness between neighbouring generated sentences as the intra-story coherence, and the relatedness between the leading context and story sentence as the intra-story relevance\footnote{This work is originally an unsurpervised metric developed for conversations. We use the same methods for evaluating our generated stories by implementing it on the sentences in a story. The code we use is from the repository \url{https://github.com/tonywenuon/dialog-coherence-metric}.}.  \textbf{Intra-story  Aggregate Metrics} i.e. repetition, coherence, and relevance, are obtained by the mean of sentence-level metrics.

% ---- 4.4.2
\subsubsection{Evaluation Results}

\paragraph{Reference metrics.}
Table~\ref{tab:referenced} shows the automatic evaluation results on both the short story dataset ROC and the long story dataset WP. 
It can be observed that EtriCA outperforms all baselines on all metrics for both datasets. Compared to the strongest baseline BART and HINT, our model reduces perplexity by $15\%$ on ROC and $25\%$ on WP, respectively. 
%, indicating the superior performance on fluency. 
For the BLEU and ROUGE metrics, EtriCA also outperforms other baselines, which demonstrates that EtriCA generates stories more closely resembling the human written reference stories. 
% In addition, the advance of EtriCA is increasing with the stories growing longer (WP longer than ROC). This indicates 
% our proposed framework more effectively incorporates context and event features, and reduces the context gap caused by the fact that when the event sequence increases the context information is relatively decreased in captured features

%the SOTA baselines on perplexity  
%for all reference metrics. EtriCA achieves the lowest perplexity, 
%\textcolor{orange}{which means it has the best predictive performance when generating utterances, so the outputs of EtriCA have better fluency.} EtriCA outperforms other baselines on both the BLEU and ROUGE scores,  \textcolor{orange}{demonstrating that EtriCA generates stories more like the human written reference stories.} 

In addition, with the ablation study, it can be observed that both the context and event features play an important role in improving the generation process. Considering the performance of \textit{- w/o leading} and \textit{- w/o events}, they indicate that  the features contained in the two kinds of inputs are complementary to each other, and both of them are essential for good story writing. Therefore, the task of how to effectively incorporate both features is important for enhancing the ability of writing a high-quality story. When EtriCA does not implement our contextualising module (abbr. cm), all the metrics substantially drop, and some metrics even become lower than those of BART$_{l+e}$ and HINT$_{l+e}$. This observation suggests that our contextualising module can more effectively fuse the heterogeneous features, and generate a richer semantic representation for the following story writing. Similarly, the sentence-level representations also improve most of the metrics, although not as much as the contextualising module. We hypothesise it is because the contextualised module has significantly reduced the gap between  event sequences and the stories (each event is paired with each sentence), making the improvement by sentence-level representations less prominent. Our hypothesis is also confirmed in the following experiments.

  % ----------- tab: unreferenced -----------
\begin{table}[tb]
\centering
\resizebox{0.95\linewidth}{!}{
\begin{tabular}{l|cc|cc}
\toprule[1pt]
\multirow{2}{*}{\textbf{Models}} & \multicolumn{2}{c}{\textbf{ROC Stories}} & \multicolumn{2}{c}{\textbf{Writing Prompts}}\\
  &  \textbf{LR-2}$ \downarrow $ & \textbf{D-4}$ \uparrow $         
  &  \textbf{LR-2}$ \downarrow $ & \textbf{D-4}$ \uparrow $  \\
\midrule
\textbf{P\&W$_{l+e}$} & 0.297 & 0.773     & 0.443 & 0.834 \\
\textbf{GPT-2$_{l+e}$} & 0.528 & 0.675    & 0.760 & 0.684 \\
\textbf{BART$_{l+e}$}  & 0.245 & \textbf{0.804}    & 0.378 & 0.894 \\

\textbf{HINT$_{l+e}$}  & 0.264 & 0.734      & \textbf{0.338} & 0.855 \\
\midrule
\textbf{EtriCA} & \textbf{0.244} & 0.799    & 0.359 & 0.889 \\ 
\textbf{- w/o sen}  & 0.286 & 0.794.     & 0.343 & \textbf{0.900} \\
\textbf{- w/o cm} & 0.245 & 0.800     & 0.514 & 0.827\\ 
\textbf{- w/o leading}  & 0.260 & 0.795      & 0.517 & 0.892\\ 
\textbf{- w/o events}  & 0.245 & 0.792   & 0.412 & 0.850 \\
\midrule
\textbf{Golden} & 0.048 & 0.906   & 0.286 & 0.950 \\
\bottomrule[1pt]
\end{tabular}
}
\caption{Automatic evaluation of unreferenced metrics on the ROC and WP datasets for the generation models writing stories conditioned on both leading text and the reference event sequence. \textbf{Golden} in the table denotes the reference stories. }
\label{tab:unreferenced}
\end{table}
% ----------- end of tab-----------

\begin{figure}[tb]
%\centering
\hspace{-0.5cm}
\includegraphics[width=9cm]{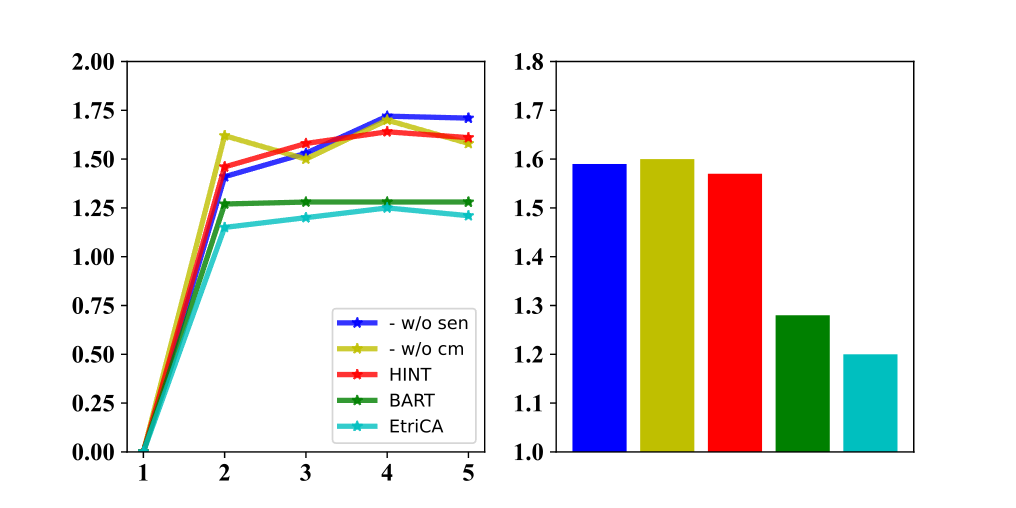}
\caption{The results of intra-story repetitions and aggregate scores on stories of the ROC dataset. The curve graphs illustrate the intra-story repetition for each sentence (leading context as the first sentence) in a story. The histograms depict the aggregate scores of intra-story repetitions over the story sentences.}
\label{fig:intra-inter-rept}
\end{figure}

\paragraph{Unreferenced Metrics.}
In another set of experiments, we examine repetition and diversity of the generated stories, where the results are reported in  Table~\ref{tab:unreferenced}. It can be observed that EtriCA gives strong performance for both lexical repetition (LR-2) and diversity (D-4), either achieves the best performance or is on a par with the best performing baseline. 
%For most metrics, EtriCA either achieves the best performance or is on a par with the best performing baseline.
%\textcolor{orange}{high performance of our approach on reference metrics} does not sacrifice  diversity. 
To further investigate how our model performs on writing along with the planned events, we follow \citet{yao2019plan} to observe the intra-story repetitions for each generated sentence, as shown in Figure~\ref{fig:intra-inter-rept}. The results show that EtriCA consistently outperforms the baselines 
for both sentence-level and story-level (i.e., aggregated) repetitious scores, 
%\textcolor{orange}{not only for these aggregate scores, but also for the whole writing process for each event}, 
indicating that EtriCA performs better  on event-triggered story writing.

% ----------- fig: intra-inter-rept -----------
% ----------- end of figure -----------

\paragraph{In-depth analysis}
Furthermore, we present more experimental results\footnote{We tried to conduct intra-story analysis (repetition, coherence, relevance) on the WP dataset. However, even in the referenced stories, there are a great number of sentences that are too short and conversational, e.g. ``well ?'' or ``you said that ?'', to be meaningfully analysed. Therefore, we only conduct intra-story experiments on the ROC dataset.} to analyse the intra-story coherence and relevance of our models. We select the two strongest baselines according to previous experimental results to compare. As shown in Table~\ref{tab:coherence_and_relevance_roc}, our approach consistently outperforms the baselines for both the intra-story coherence and relevance. This indicates our contextualising module improves the feature capturing for both the context and event features to enhance the logical relatedness between story sentences, and between the story and the input.

% ----------- tab: coherence_and_relevance_roc ----------
\begin{table}[tb]
\centering
\resizebox{\linewidth}{!}{
\begin{tabular}{l|ccc|ccc}
\toprule[1pt]
\multirow{2}{*}{\textbf{Models}} & \multicolumn{3}{c}{\textbf{Coherence}} & \multicolumn{3}{c}{\textbf{Relevance}} \\
  & \textbf{wiki.} & \textbf{twit.}& \textbf{comm.}  & \textbf{wiki.} & \textbf{twit.}& \textbf{comm.}  \\
\midrule
\textbf{BART$_{l+e}$}  & 0.4658 & 0.6293 & 0.5865      & 0.5316 & 0.6710 & 0.6439    \\
\textbf{HINT$_{l+e}$}  & 0.4627 & 0.6276 & 0.5818      & 0.5323 & 0.6718 & 0.6427  \\
\midrule
\textbf{EtriCA} & 0.4667 & 0.6306 & \textbf{0.5876}     & 0.5332 & 0.6722 & 0.6445  \\ 
\textbf{- w/o sen} & \textbf{0.4680} & \textbf{0.6322} & 0.5864     & \textbf{0.5356} & \textbf{0.6745} & \textbf{0.6457}    \\
\textbf{- w/o cm} & 0.4602 & 0.6232 & 0.5775       & 0.5281 & 0.6676 & 0.6381   \\ 
\midrule
\textbf{Golden} & 0.6631 & 0.7996 & 0.8298        & 0.6610 & 0.7997 & 0.8265    \\ 
\bottomrule[1pt]
\end{tabular}
}
\caption{The results of aggregate scores of intra-story coherence and relevance for the ROC dataset, which are calculated based on semantic embeddings. \textbf{wiki.}, \textbf{twit.}, \textbf{comm.} denotes the glove embeddings of ``Wikipedia 2014 + Gigaword 5 (6B tokens)'', ``Twitter (2B tweets, 27B tokens)'', and ``Common Crawl (42B tokens)'', respectively.}
\label{tab:coherence_and_relevance_roc}
\end{table}
% ----------- end of tab-----------

Additionally, the ablation results, in which EtriCA and ``- w/o sen'' have very close performance, also confirm the aforementioned hypothesis that the feature capturing mechanism of the contextualising module partly replaces the functions of sentence-level representations. 

% ----------- fig: sent-coh-rel -----------
\begin{figure*}[tb]
\centering
\hspace{-0.2cm}\includegraphics[width=0.7\linewidth]{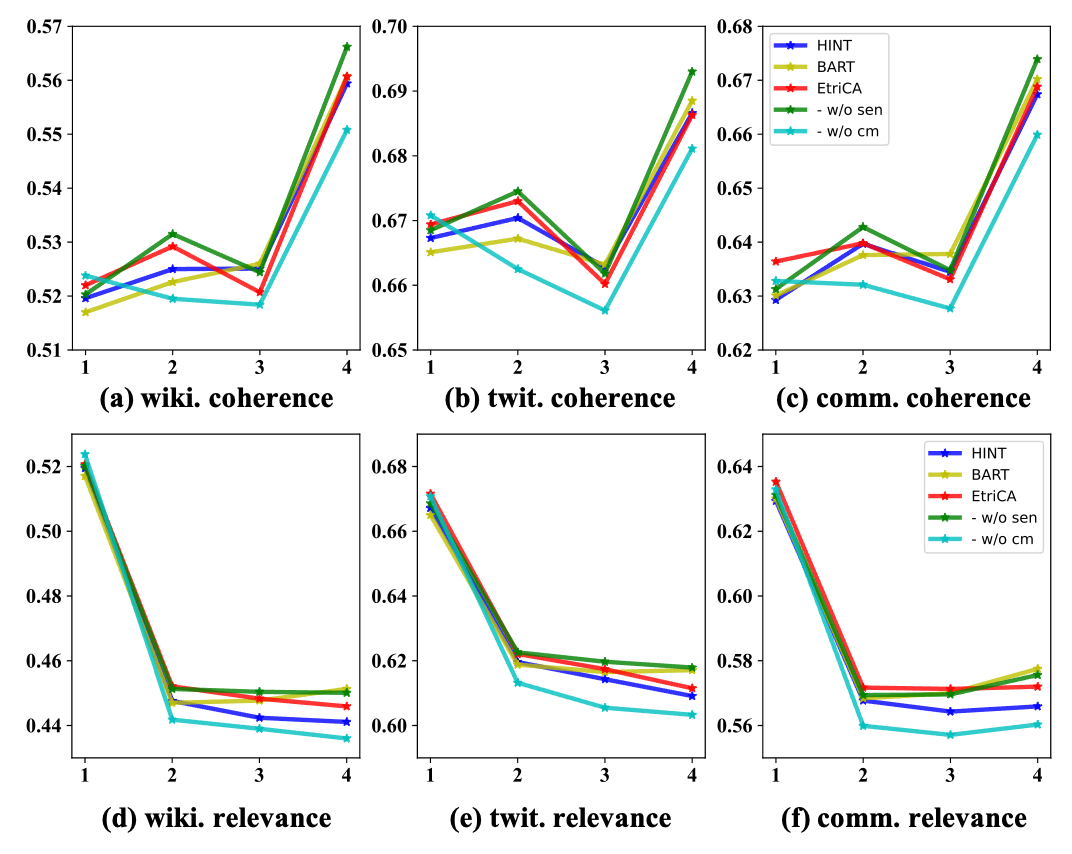}
\caption{The results of intra-story coherence and relevance on the ROC dataset.}
\label{fig:sent-coh-rel}
\end{figure*}
% ----------- end of figure -----------

Figure~\ref{fig:sent-coh-rel} further shows that our model significantly outperforms baselines for coherence and relevance, which indicates its advance in feature capturing and generating more event and context related stories.

% ----------- tab: simple case study -----------
\begin{table}[ht]
\centering
\resizebox{\linewidth}{!}{
\begin{tabular}{{p{1.2\linewidth}}}
\toprule[2pt]
\textbf{Leading Context:} [MALE] had lost his dog over a month ago . \\
\textbf{Event Sequence:} missed dog $\rightarrow$ notices something $\rightarrow$ sees dog $\rightarrow$ turns out be \\
\midrule[1pt]
\textbf{BART$_{l+e}$:} He \textbf{missed} his \textbf{dog} for a whole month. One day he \textbf{notices something} moving and is startled. He \textbf{sees} the \textbf{dog} on the floor. It \textbf{turns out} to \textbf{be} a squirrel. \\
\midrule
\textbf{HINT$_{l+e}$:} One day [MALE] \textbf{missed} his \textbf{dog}. He \textbf{notices something} about her name on the dog's tag. [MALE] \textbf{sees} the \textbf{dog} in the tags. It \textbf{turns out} it could \textbf{be} a dog from the police department.  \\
\midrule[1pt]
\textbf{EtriCA:} He \textbf{missed} his \textbf{dog} badly. He \textbf{notices something} strange on the curb. He \textbf{sees} the \textbf{dog} outside. It \textbf{turns out} to \textbf{be} a stray dog.  \\
\midrule
\textbf{- w/o sen: }He had \textbf{missed} his \textbf{dog} so much that he had to search for him. As he was searching, he \textbf{notices something} about a dog. He \textbf{sees} the \textbf{dog} with a bag. It \textbf{turns out} to \textbf{be} a stray, a wad of dog spray.  \\
\midrule
\textbf{- w/o cm: }[MALE] \textbf{missed} his \textbf{dog} this summer. He \textbf{notices something} on his neighbor's wall about a house. [MALE] notices the dog was very sad. it \textbf{turns out} that there must \textbf{be} a really sad day next time. \\ 
\bottomrule[2pt]
\end{tabular}
}
\caption{A case study for an example in \textbf{ROC Stories}. \textit{[MALE]}, \textit{[FEMALE]}, and \textit{[NEUTRAL]} are the specital tokens to replace names in stories. The highlighted bold words denote the events corresponding to the given event sequence.}
\label{tab:simple_case_study}
\end{table}
% ----------- end of tab-----------

% ######################### Section 5.2 ########################
\subsection{Human Evaluation} \label{human-evaluation}

We conducted human evaluation based on pair-wise comparisons with two competitive baselines and the ablated model without our proposed contextualising module. We randomly sample 150 stories from the test datasets of ROC Stories\footnote{WP is not used to conduct experiments since long stories are difficult to get acceptable annotation agreement.}. There are 3 evaluators 
% with full proficiency in English
invited to choose which generated story is better (Win/Lose/Tie) on three aspects: (\romannumeral1) \textbf{Fluency}  considers each sentence in isolation and measures the quality from a linguistic perspective, e.g., the grammatical correctness or the correct representation of semantic meaning; (\romannumeral2) \textbf{Coherence}  measures the logical relatedness between  story sentences; (\romannumeral3) \textbf{Relevance} measures the context relevance between stories and the leading contexts. When summarising the human annotation result, the final results are counted by majority voting.

% ----------- tab: human evaluation-----------
\begin{table}[tb]
\centering
\resizebox{0.85\linewidth}{!}{
\begin{tabular}{r|lc|c}
\toprule[1pt]
\multirow{2}{*}{\textbf{Choices(\%)}} & \multicolumn{3}{c}{\textbf{Etri.} vs. \textbf{\textit{w/o cm}}}  \\
\cline{2-4} 
& \textbf{Etri.} & \textbf{\textit{w/o cm}} & \textbf{Kappa}   \\
\midrule
\textbf{Fluency}  & \textbf{36.1}$^{**}$ & 18.0 & 55.3  \\
\textbf{Coherence}  & \textbf{40.2}$^{**}$ & 22.7 & 48.9 \\
\textbf{Relevance}  & \textbf{23.2} & 21.7 & 48.5  \\

\midrule[1pt]
\multirow{2}{*}{\textbf{Choices(\%)}} & \multicolumn{3}{c}{\textbf{Etri.} vs. \textbf{BART$_{l+e}$}} \\
\cline{2-4} 
& \textbf{Etri.} & \textbf{BART$_{l+e}$} & \textbf{Kappa} \\
\midrule
\textbf{Fluency}  & \textbf{33.6}$^{**}$ & 16.4 & 56.2  \\
\textbf{Coherence}  & \textbf{32.8}$^{*}$ & 19.1 & 48.1  \\
\textbf{Relevance}  & \textbf{16.8} & 9.8 & 45.1  \\

\midrule[1pt]
\multirow{2}{*}{\textbf{Choices(\%)}} & \multicolumn{3}{c}{\textbf{Etri.} vs. \textbf{HINT$_{l+e}$}} \\
\cline{2-4} 
& \textbf{Etri.} & \textbf{HINT$_{l+e}$} & \textbf{Kappa} \\
\midrule
\textbf{Fluency}   & \textbf{32.3}$^{**}$ & 17.3 & 55.4 \\
\textbf{Coherence}   & \textbf{35.3}$^{*}$ & 21.8 & 58.3 \\
\textbf{Relevance}  & \textbf{14.9} & 8.5 & 50.9 \\
\bottomrule[1pt]
\end{tabular}
}
\caption{Human evaluation results on the ROC dataset. The scores stand for the percentage of model chosen in pair comparisons (win another model). Kappa means the Fleiss' Kappa \cite{fleiss1971measuring} coefficient that is used to measure inter-annotator agreement. All of our results have reached moderate agreement. $*$ refers to significance at p<0.05, whilst $**$ refers to significance at p<0.01, on a sign test.}
\label{tab:human_eval}
\end{table}
% ----------- end of tab-----------

As shown in Table~\ref{tab:human_eval}, EtriCA outperforms SOTA baselines in terms of fluency, coherence, and relevance. All generation models have relatively little conflict with the given input, so they all have good performance on relevance, causing less differences in relevance. On the other hand, the advances on fluency and coherence are very significant, indicating the advantage of our contextualising module in capturing high-level features from the context and event sequences.

% ######################### Section 5.3 ########################
\subsection{Case Study} \label{sec:case-study}
As is shown in \autoref{tab:simple_case_study}, examples indicate that compared to baseline models, EtriCA generates a more logically related story with given inputs, and contains fewer confusing expressions inside the story. (More examples in Appendix.~\ref{apx:case-study})

% =============================== Section 6 ==================================
\section{Conclusion}
We present a controllable story generation task conditioned with leading contexts and event sequences. We also provide two new datasets with extra annotated events, and introduce a new set of automatic metrics to measure coherence and fluency. We propose EtriCA, a novel generation model, which better exploits context and event features with a cross attention based contextualising network. Through extensive experiments and analyses, we demonstrate EtriCA can generate stories with better fluency, coherence, and relevance compared to competitive baselines.

% =============================== Section 7 ==================================
\section*{Acknowledgements}
Chen Tang is supported by the China Scholarship Council (CSC) for his doctoral study (File No.202006120039). We also gratefully acknowledge the anonymous reviewers for their insightful comments.

% =============================== Section 8 ==================================
\section*{Limitations}
According to our methodology and some observations from experiments, we conclude the limitations around our study as follows:
\begin{itemize}[noitemsep,nolistsep,leftmargin=*]
    \item \textbf{Robustness:} The quality of given events may affect the robustness of generation models. It is very hard to define whether a plot (an event sequence) is ``interesting'' or ``bad''.  When provided with unusual or strange events for a story, the generation model struggles to follow the events cohesively.
    \item \textbf{Sequence Length:} Our neural models cannot write stories that are too long in length. Because of the features of neural encoders and decoders, the inputs and outputs generally have a certain length limitation, e.g. 1024 tokens.
    \item \textbf{Generalisation:} The training datasets of stories limit the topics of open domain story writing. Although large-scale pre-training has been widely adopted, our neural models still struggle to tackle some topics, writing styles, etc. For instance, the performance of story generation on WP is relatively worse than on ROC, because WP contains  spoken language and irregular expressions.
    \item \textbf{Experiments:} For the limitation of resources, we did not conduct some further experiments in this study. For instance, we did not further study the performance of difference event representations because it is not the focus of this study.
\end{itemize} 

% =============================== Reference ==================================
\bibliography{bibs/Sec1-introduction,
bibs/Sec2-related-work,
bibs/Sec3-methodology,
bibs/Sec4-exp-setup,
bibs/Sec5-exp-result,
bibs/Appendix}
\bibliographystyle{acl_natbib}

% =============================== Appendix ==================================
\appendix
\section{Appendix}
\label{sec:appendix}

% ######################### Section A.1 ########################
\subsection{Implementation Details} \label{apx:implementation}
\paragraph{Hyparameters of model} To leverage pre-trained parameters, most of the hyparameters for the encoder, decoder, and embedding layers are restored from the public checkpoints\footnote{BART from the bart-base \url{https://huggingface.co/facebook/bart-base}, and GPT-2 from  gpt2-base \url{https://huggingface.co/gpt2}} on Huggingface. In addition, we set other parameters as follows: The residual scale factor $\beta$ in Equation.~\ref{eq:beta} is set to 0.1. The margin $\Delta$ in Equation.~\ref{eq:delta} is set to 0.1. The scale factor $\lambda$ in Equation.~\ref{eq:lam} is set to 0.1. There are also some parameters are learnable via training on datasets. In our framework, the two encoders and the decoder all have 6 hidden layers implementing 12-head attention mechanism. Both encoders and decoders share the embeddings layer containing vocabulary up to 50,625 tokens
with Byte-Pair Encoding \cite{radford2019language}, and additional special tokens mentioned in Sec.~\ref{sec:methodology}.

\paragraph{Training Settings} 
Our experiments are carried out on multiple GPUs (e.g., RTX A4000) on a cloud platform, so for the convenience of reproduction, we fix the random seed to 42. 
When training neural models, we implement PyTorch Lightning\footnote{It offers a lot of api interfaces which simplify engineering. \url{https://www.pytorchlightning.ai/}} framework to set up training processes. The training parameters setup are listed as follows: The \textit{batch size} is set to 64; The \textit{learning rate} is 8e-5; \textit{max source length} is set to 1024; The optimiser uses Adam \cite{kingma2014adam}, and the $\epsilon$ of Adam is set to 1e-8. The whole training process will last for 5 \textit{epochs}, but the result only considers the checkpoint with the best performance on the metric of \textit{loss} (the lowest). It is worth mentioning that EtriCA needs two separate encoders to encode context (natural language) and events (concatenated serialised events) separately, but the encoder of public BART checkpoint is only pre-trained on natural language text. Therefore, to make the event encoder learn event features better, we firstly train a BART model on stories given both the context and planned events, and  then restore the pre-trained encoder paratermeters to the event encoder of EtriCA.

\paragraph{Inference Settings} When evaluating and testing, we adopt the nucleus sampling \cite{holtzman2019curious} strategy to generate texts. We also change the \textit{batch size} to 15 when doing the inference, because nucleus sampling requires large amount of memory,

% ######################### Section A.2 ########################
\subsection{Details of Event Schema} \label{apx:schema}

\begin{figure}[htpb]
\centering
\includegraphics[width=\columnwidth]{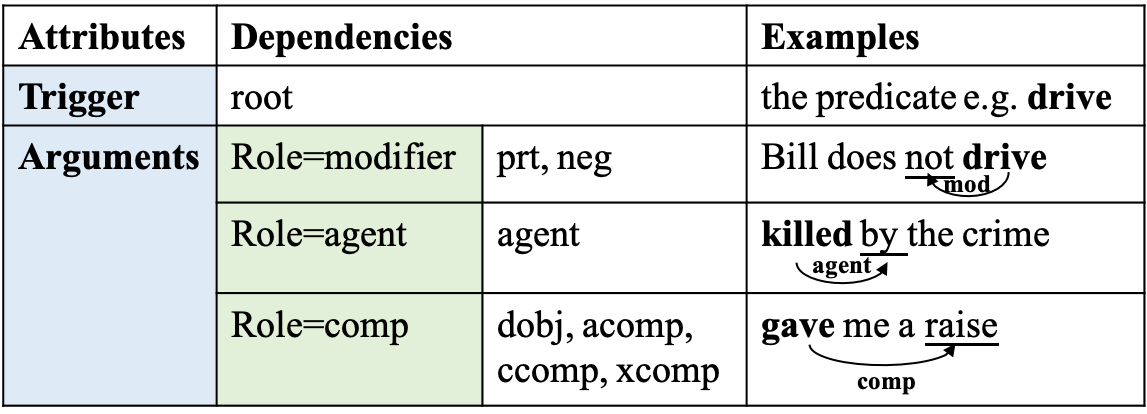}
\caption{The schema of event shows the relations with event arguments and word dependencies. We offer some examples to indicate these dependencies, e.g., in "Bill does not drive" "not" is a negation (\textbf{neg}) to "drive", so it is an event modifier. }
\label{fig:schema}
\end{figure}

An event is supposed to represent an important change that happens, so generally the representation of an action. The schema for an event aims to include all relevant roles to the action and filter trivial details for representation. Inspired by the work of \citet{rusu-etal-2014-unsupervised,bjorne-salakoski-2018-biomedical} which used dependency parser to capture dependencies between words belonging to different
clauses, we extract event mentions from sentences according to the hierarchy of typed dependencies \citep{de2008stanford} (see details in Appendix.~\ref{apx:schema}). In this way we can obtain more informative and unambiguous events compared to single-verb events used in previous work \cite{jhamtani-berg-kirkpatrick-2020-narrative, guan-etal-2020-knowledge, kong2021stylized}. The schema is shown in Figure \ref{fig:schema}. 

% ----------- fig:schema -----------
% ----------- end of fig -----------

As shown in Figure.~\ref{fig:schema}, event arguments are extracted according to selected dependencies between words. Here, we give the details of these dependencies, and Table.~\ref{tab:schema_details} indicate the roles of these dependencies in a sentence. (more details of dependencies see \citet{de2008stanford})
% ----------- tab: details-of-Event-Schema -----------
\begin{table}[htpb]
\centering
\resizebox{\linewidth}{!}{
\begin{tabular}{lcc}
\toprule[1pt]
\textbf{Dep.} & \textbf{Full Name} & \textbf{Example} \\
\hline

 \textbf{prt} & phrasal verb particle & [shut]-\textit{prt}->[down] \\
 \textbf{neg} & negation modifier & [drive]-\textit{neg}->[not] \\
 \textbf{agent} & agent & [killed]-\textit{agent}->[police] \\
 \textbf{dobj} & direct object & [gave]-\textit{dobj}->[raise] \\
 \textbf{acomp} & adjectival complement & [looks]-\textit{acomp}->[beautiful] \\
 \textbf{ccomp} & clausal complement & [says]-\textit{comp}->[like] \\
 \textbf{xcomp} & open clausal complement & [like]-\textit{xcomp}->[swim] \\
 
\bottomrule[1pt]
\end{tabular}
}
\caption{Details of dependencies in Event Schema. Examples are extracted with the format of [head]-\textit{dependency}->[tail].}
\label{tab:schema_details}
\end{table}
% ----------- end of tab -----------

The schemas of events are required to consider performance with respect to both generalisation and representation. More dependencies included can make an event more informative but may cut down its generalisation ability. For instance, \textit{Subject} (e.g. I, you, Kent, ...) is useful to state the character of an event, but stories usually have different characters that may make it difficult for extracted events from one story to be reused in another story. E.g., "Kent is driving" and "He is driving" are the same meaning but if the subject "Kent" is extracted as an event role, it is very hard to predict same event for another story, which means the generalisation is damaged. According to similar criterion we select key roles as the arguments to events with the consideration of both generalisation and representation.

% ----------- fig:survey -----------
\begin{figure*}[ht]
\centering
\includegraphics[width=2\columnwidth]{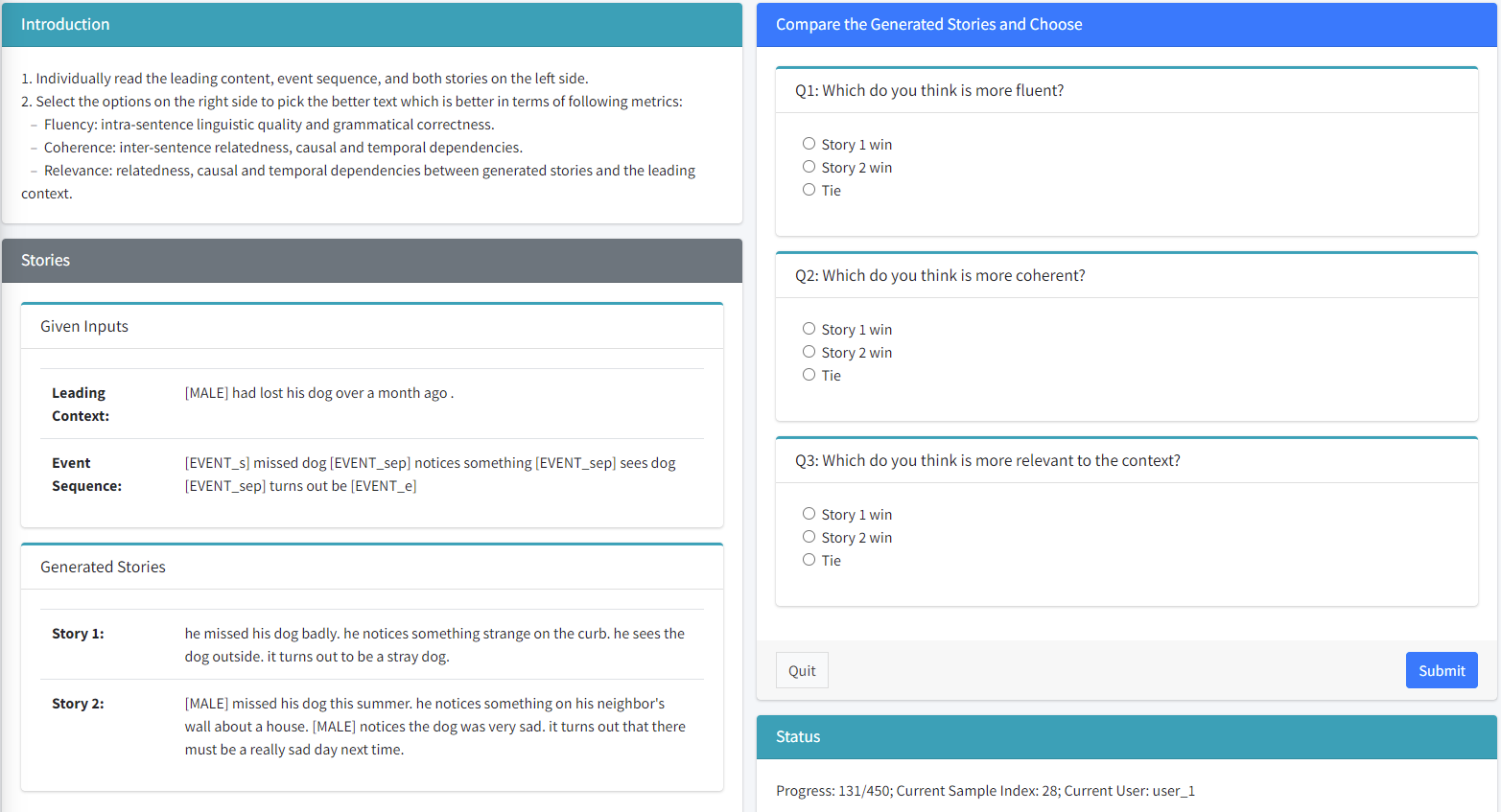}
\caption{The snapshot of our evaluation interface. The stories are randomly collected from the \textbf{ROC} dataset, and the annotators are required to choose a choice for each question on the right colomn. For the convenience of accurate annotation, the system allow annotators to directly compare different generated stories with given input. A survey will be automatically recorded with all the three questions answered and the "submit" button pressed.}
\label{fig:survey}
\end{figure*}
% ----------- end of fig -----------

% ######################### Section A.3 ########################
\subsection{Details of Event Extraction} \label{apx:planning}
We extract events from the text of training dataset including reference stories and leading contexts. The data structure of an event is a set including relevant trigger and arguments in a sentence. We firstly use $\mathit{spaCy}$\footnote{\url{https://spacy.io/}} to parse dependencies between words in a sentence, and then annotate event trigger and arguments according to the dependencies. An event  $ e $ contains attributes introduced in Figure \ref{fig:schema}, in which the event trigger as the root is generally the predicate. Before encoders accept text as the input, extracted events will be serialised to text format to feed the model.

Since existing story datasets do not have the reference storylines paired with reference stories, we develop an event extractor to extract event sequences from reference stories as the storylines. We follow the route to represent events as verb phrases. Verbs as the anchor of sentences can be seen as the \textit{event trigger}, so our primary goal is to extract all the key roles (as \textit{event arguments}) related to the event trigger. The neighbourhood of extracted events will be considered as temporal relations. 

With temporally related events from training stories, we construct an event graph denoted as $ G $, which is an isomorphic graph with single event type and single relation type. We suppose $ G $ is a data structure composed by triples in $e_h, r, e_t$ format. The workflow of the extraction process is explained as follows:

% ----------- algorithm:extraction -----------
\begin{algorithm}[!ht]
\DontPrintSemicolon
  \KwInput{A story $ S $ with $ m $ sentences}
  \KwOutput{Event Sequence $ E $ for $ S $  containing $ m $ event objects}
  Initialise $ E \leftarrow \varnothing $ and $ roles \leftarrow \{\mathit{trigger}, \mathit{mod}, \mathit{agent}, \mathit{comp}\} $
  \ForEach{$s^i$ in $S$}{
        Initialise $ e_i \leftarrow \varnothing $
        
        Normalise $ s^i $ and get dependencies $\mathit{dep}_i$ with $\mathit{spaCy}$
        
        Extract event trigger $t$ and position $p_t$ from $\mathit{dep}_i$
        
        $ e_i.\mathit{trigger} \leftarrow t$
        
          \ForEach{$\mathit{role}$ in $\mathit{role}$}{
              \If{$t \in \mathit{dep}_i.heads$ and $\mathit{role} \in \mathit{dep}_i.tails $}{
                  Extract $(\mathit{role}, p_{\mathit{r}})$ from $\mathit{dep}_i$
            
                  $ e_i.\mathit{role} \leftarrow (\mathit{role}, p_{\mathit{r}}) $
              }
      }
      $ e_i.\mathit{string} \leftarrow r \in \mathit{roles} $ aligned by $ p_r \uparrow$
      
      $E$ append $e_i$
  }
\caption{Extract Event Sequence $E$\label{alg:extract}}
\end{algorithm}
% ----------- end of algorithm-----------

% ----------- tab: case study -----------
\begin{table*}[ht]
\centering
\resizebox{130mm}{!}{
\begin{tabular}{l|ll}
\toprule[2pt]
 \multirow{2}{*}{\textbf{Input}} & \textbf{Leading Context} & [MALE] had lost his dog over a month ago .\\
& \textbf{Event Sequence} & missed dog $\rightarrow$ notices something $\rightarrow$ sees dog $\rightarrow$ turns out be \\
\midrule[1pt]
\textbf{P\&W$_{l+e}$} & \multicolumn{2}{p{0.9\textwidth}}{He wished he could live with his friend. He 'd run in them all the time. But one day, he woke up exhausted. He went to the doctor with his best friend.} \\
\midrule
\textbf{GPT-2$_{l+e}$} & \multicolumn{2}{p{0.9\textwidth}}{[MALE] was only a parent at the time. [MALE] \textbf{notices} the dog and he lets it go. He notices the dog has been moved and so he notices what happened. [MALE] then realises that it is a bad dog and there is something wrong with his life.} \\
\midrule
\textbf{BART$_{l+e}$}  & \multicolumn{2}{p{0.9\textwidth}}{He \textbf{missed} his \textbf{dog} for a whole month. One day he \textbf{notices something} moving and is startled. He \textbf{sees} the \textbf{dog} on the floor. It \textbf{turns out} to \textbf{be} a squirrel.} \\
\midrule
\textbf{HINT$_{l+e}$}  & \multicolumn{2}{p{0.9\textwidth}}{One day [MALE] \textbf{missed} his \textbf{dog}. He \textbf{notices something} about her name on the dog's tag. [MALE] \textbf{sees} the \textbf{dog} in the tags. It \textbf{turns out} it could \textbf{be} a dog from the police department.}  \\
\midrule\midrule
\textbf{EtriCA} & \multicolumn{2}{p{0.9\textwidth}}{He \textbf{missed} his \textbf{dog} badly. He \textbf{notices something} strange on the curb. He \textbf{sees} the \textbf{dog} outside. It \textbf{turns out} to \textbf{be} a stray dog.}  \\
\midrule
\textbf{- w/o sen} & \multicolumn{2}{p{0.9\textwidth}}{He had \textbf{missed} his \textbf{dog} so much that he had to search for him. As he was searching, he \textbf{notices something} about a dog. He \textbf{sees} the \textbf{dog} with a bag. It \textbf{turns out} to \textbf{be} a stray, a wad of dog spray.}  \\
\midrule
\textbf{- w/o cm} & \multicolumn{2}{p{0.9\textwidth}}{[MALE] \textbf{missed} his \textbf{dog} this summer. He \textbf{notices something} on his neighbor's wall about a house. [MALE] notices the dog was very sad. it \textbf{turns out} that there must \textbf{be} a really sad day next time.} \\ 
\midrule
\textbf{- w/o leading} & \multicolumn{2}{p{0.9\textwidth}}{He \textbf{missed} his \textbf{dog}. [MALE] \textbf{notices something} in the area. he \textbf{sees} a \textbf{dog}. it \textbf{turns out} to \textbf{be} a black dog.} \\ 
\midrule
\textbf{- w/o events} & \multicolumn{2}{p{0.9\textwidth}}{He was devastated by the loss. He decided to pull a long string of nail polish. He found a couple of old nail polish cans that were very old. His dog enjoyed his touches.}\\
\bottomrule[2pt]
\end{tabular}
}
\caption{A case study of generated stories conditioned with a leading context and an event sequence collected from \textbf{ROC Stories}. \textit{[MALE]}, \textit{[FEMALE]}, and \textit{[NEUTRAL]} are the specital tokens to replace names in story. The highlighted bold words denote the events corresponding to the given event sequence.}
\label{tab:case_study}
\end{table*}
% ----------- end of tab-----------

% ######################### Section A.4 ########################
\subsection{Details of Methodology} \label{apx:methodology}
\paragraph{Sentence Prediction Task} 
An auto-regressive decoder will predict $y_t$ based on prior tokens $y_{<t}$, so we let a neural model learn to generate a special hidden state $H_{sep}^i$ at the position of special token $[sep_i]$ where $i$ denotes the $i$-th sentence. We use \textit{Sentence-Bert} to obtain a numeric vector $F_i^{sent}$, which contains the features of sentences through representation learning. Then we force the similarity score $sim_{ij}^{s}$ between generated sentences to fit similarities between sentences in the reference stories. The calculation of similarity score is shown below:
% ----------- equation-----------
\begin{align}
    & F_i^{sent} = \textit{Sentence-Bert}(\{s_1^i, ..., s_n^i\}) \\
    & sim_{ij}^{s} = cosine(F_i^{sent}, F_j^{sent}) \\
    & u_{ij} = (H_{sep}^i)^\intercal W^{sep} H_{sep}^j \\
    & sim_{ij}^{y} = sigmoid(u_{ij} + u_{ji}) 
\end{align}
% ----------- end of equation -----------
where $i$ and $j$ denote the indices of a sentence. $sim$ denotes the similarity. $sim_{ij}^{s}$, the ground-truth similarity, is computed by $cosine$ similarity between outputs of \textit{Sentence-Bert}. $u_{ij}$ is an intermediate variable of similarity obtained from predicted sentence representations, and $W^{sep} $ denotes a trainable parameter. To guarantee $sim_{ij}^{y}$ is symmetric with respect to either $i$ to $j$ or $j$ to $i$, $u_{ij}$ and $u_{ji}$ are both incorporated. 

\paragraph{Contextualised Features Representation} \label{apx:contextualised}
Conventional end-to-end models usually concatenate embeddings of different input, e.g. here we concatenate $C$ and $E$ as the input to the baseline model. The encoders such as LSTM, Bert-like encoders will incorporate the heterogeneous features.
However, there may be some potential problems: (\romannumeral1) $C$ and $E$ have different word distribution, since $C$ is natural language but $E$ is not. A single encoder may not capture the two type of features in a single input numeric vector, efficiently and effectively; (\romannumeral2) With the increment of the length of stories, the relative size of $E$ will surpass $C$. The single encoder may pay less attention to $C$ features, e.g. the vectors $C$ involved in the calculations of attention scores become relatively less when the plan events increase.

% ######################### Section A.5 ########################
\subsection{Details of Human Evaluation}
We developed an evaluation system, which helps us to collect annotations of evaluation, to make the story pairs for annotation anonymous, fairly shuffled, and easy to compare. Figure~
\ref{fig:survey} is the snapshot of our annotating process.

Evaluators are required to follow the annotation standards shown on the left top corner. Considering the different biases among individuals, we also notify every annotator our standards set for this task: (\romannumeral1) \textbf{Fluency}  considers the errors shown in generated text, e.g. grammatical errors $\geq$ spelling errors $\geq$ unnatural repetitions \textgreater language quality. (\romannumeral2) \textbf{Coherence}  focuses on the logical relatedness between sentences. We asked annotators to count all the incoherent parts, and consider how many word edits would be needed to make the story coherent (i.e. fewer edits needed = more coherent story). (\romannumeral3) \textbf{
Relevance} focuses on the relatedness between generated sentences  and the leading context. However, it is very subjective to judge if a story is "interesting" or not relevant. Therefore, we suggest evaluators to judge how irrelevant a story is by counting the conflict generated sentences with the leading context.

% ######################### Section A.6 ########################
\subsection{Case Study} \label{apx:case-study}
As is shown in Table~\ref{apx:case-study}, EtriCA can generate better stories considering both context relatedness and story quality. 
%The baseline models usually have more non-fluent problems, e.g. the sentence "He notices the dog has been moved and so he notices what happened." as a complete sentence is very confusing to read. We consider this is a bad performance of language quality (fluency). 
The strong baselines models, i.e. \textbf{BART$_{l+e}$} and \textbf{HINT$_{l+e}$}, generate stories with good obedience to the planned event sequences, and relatively good fluency. However, they fail to write  reasonable stories with the logical relatedness to the ongoing circumstances. For instance,  "It turns out to be a squirrel." is a good sentence and also uses the event "turns out be", but it has nothing to do with the topic - "the dog is missing", and no coherence with previous sentences as well.

In terms of the results obtained for the ablation study, we see the importance of different components to the whole generation model. If there is no planned event sequence (see \textbf{- w/o events}), it is very hard to let a neural model write a coherent story, which is also demonstrated in previous work \cite{yao2019plan}. If there is no given leading context (see \textbf{- w/o leading}), neural models struggle to unfold the planned events, because the neural model does not understand the concept of a "topic" for a story,  it may cause confusion. Without the contextualising module (see \textbf{- w/o cm}), the neural model struggles to process the heterogeneous features from context and events.

\end{document}